\documentclass[9pt, conference]{IEEEtran}
\IEEEoverridecommandlockouts
   \usepackage{setspace}
\usepackage{lineno}
\usepackage{wrapfig}

\usepackage{placeins}   % for FloatBarrier

\usepackage{graphicx}
\usepackage{floatrow}
\usepackage{sidecap}
\usepackage{caption}
\usepackage{calc} % for caption positioning

\usepackage{algorithm}

\usepackage{algorithm}
\usepackage{algorithmicx}
\usepackage{algpseudocode}
\usepackage{amsmath}
\usepackage{amssymb}
\usepackage{bm} % For bold math symbols

\usepackage{enumitem}
\usepackage{changepage}

   \usepackage[colorlinks=true,
               linkcolor=red,       % Color for internal links (sections, pages, etc.)
               citecolor=green,      % Color for citations
               urlcolor=blue,         % Color for URLs
               filecolor=magenta     % Color for files
               ]{hyperref}

\usepackage{natbib}
\usepackage{xcolor}         % colors
\usepackage[utf8]{inputenc} % allow utf-8 input
\usepackage[T1]{fontenc}    % use 8-bit T1 fonts
\usepackage{url}            % simple URL typesetting
\usepackage{booktabs}       % professional-quality tables
\usepackage{tablefootnote}
\usepackage{amsfonts}       % blackboard math symbols
\usepackage{nicefrac}       % compact symbols for 1/2, etc.
\usepackage{microtype}      % microtypography

\definecolor{myblue}{rgb}{0.255, 0.412, 0.882} % R, G, B values

\definecolor{green1}{RGB}{10,158,10}
\definecolor{blue1}{RGB}{17,85,204}
\definecolor{red1}{RGB}{204,0,0}
\definecolor{mygray}{gray}{0.85}

\definecolor{myblue2}{RGB}{232, 241, 247} % R, G, B values

\usepackage{url}            % simple URL typesetting
\usepackage{booktabs}       % professional-quality tables
\usepackage{amsfonts}       % blackboard math symbols
\usepackage{nicefrac}       % compact symbols for 1/2, etc.
\usepackage{microtype}      % microtypography
\usepackage{lipsum}
\usepackage{wrapfig}
\usepackage{siunitx}
\usepackage{sidecap}

\usepackage{amsmath,amsfonts,amssymb,amsthm}
\usepackage{mathtools}
\usepackage{multirow}

\usepackage{bbm}
\usepackage{colortbl}
\usepackage{booktabs}
\usepackage{multirow}
\usepackage{siunitx}
\usepackage{array}

\usepackage{subcaption}

\renewcommand{\mathbf}[1]{\boldsymbol{\mathit{#1}}}

   \usepackage[colorlinks=true,
               linkcolor=red,       % Color for internal links (sections, pages, etc.)
               citecolor=green,      % Color for citations
               urlcolor=blue,         % Color for URLs
               filecolor=magenta     % Color for files
               ]{hyperref}

\usepackage{enumitem}

\newcolumntype{P}[1]{>{\hspace{1ex}}p{#1}<{\hspace{1ex}}}

\newtheorem{theorem}{Theorem}
\newtheorem{corollary}{Corollary}[theorem]

\newtheorem{prop}[theorem]{Proposition}

\newtheorem{property}[theorem]{Property}

\begin{document}

\title{A Family of Kernelized Matrix Costs for \\Multiple-Output Mixture Neural Networks\vspace{-10pt}}
% {\footnotesize \textsuperscript{*}Note: Sub-titles are not captured for https://ieeexplore.ieee.org  and

\author{%
  Bo Hu, \; Jos{\'e}~C.~Pr{\'\i}ncipe\\
  Department of Electrical and Computer Engineering\\
  University of Florida\\
  \texttt{hubo@ufl.edu\; principe@cnel.ufl.edu }\\\vspace{-15pt}}

\maketitle

\begin{abstract}
Pairwise distance-based costs are crucial for self-supervised and contrastive feature learning. Mixture Density Networks (MDNs) are a widely used approach for generative models and density approximation, using neural networks to produce multiple centers that define a Gaussian mixture. By combining MDNs with contrastive costs, this paper proposes data density approximation using four types of kernelized matrix costs in the Hilbert space: the scalar cost, the vector-matrix cost, the matrix-matrix cost (the trace of Schur complement), and the SVD cost (the nuclear norm), for learning multiple centers required to define a mixture density.
\end{abstract}

\vspace{-5pt}

\section{Introduction}\vspace{-3pt}

Costs that utilize pairwise distances, whether exponential or $L_2$, are central to self-supervised and contrastive feature learning, including MoCo~\cite{he2020momentum}, SimCLR~\cite{chen2020simple}, Barlow Twins~\cite{grill2020bootstrap}, SimSiam~\cite{chen2021exploring}, VICReg~\cite{wang2021vicreg}, VICRegL~\cite{bardes2022vicregl}, FastSiam~\cite{pototzky2022fastsiam}. These costs often include a minimization term for the intra-class invariance, and a maximization term for inter-class diversity. Similar to eigendecomposition, the minimization is similar to finding an invariant equilibrium for a linear operator; the maximization is similar to enforcing orthonormality. As shown in our previous work~\cite{hu2022cross, ma2024learning, hu2023feature, hu2024learning}, such cost structures can be viewed as decomposing a linear operator of a density ratio for dependence measurement.

A prevalent approach to generative models and density approximation is fitting data with a Gaussian mixture, using a neural network capable of producing multiple centers to define the mixture, known as Mixture Density Networks (MDNs)~\cite{bishop1994mixture}. One can spot the analogy here: like defining multi-dimensional feature vectors earlier, the network here defines multiple centers for a mixture. Suppose the data density is $p(X)$ and the model is $q(X)= \int q(c)w(c)\mathcal{N}(X-m(c);v(c)) dc$, a Gaussian mixture parameterized by a neural network. We have identified at least four ways to define contrastive costs to approximate $p$ with $q$:
\begin{enumerate}[leftmargin=*]
\item Using the Cauchy-Schwarz inequality, we directly define the inner product normalized by the norm $\frac{\langle p, q \rangle^2}{\langle q ,q\rangle}$ as a scalar cost, upper bounded by the norm of the data density $\langle p, p\rangle$. Maximizing this ratio makes the bound tight and $q=p$.
\item Inspired by how the linear least-square solution is the famous $\mathbf{R}^{-1}\mathbf{P}$, we view $q(X)$ as a series of Gaussian residuals $q_1, q_2,\cdots,q_K$. Each $q_k(X) = \mathcal{N}(X-m_k;v_k)$ is a single Gaussian. Optimal weights for predicting the data density $p(X)$ using these residuals should be given by $\mathbf{R}^{-1}\mathbf{P}$, where $\mathbf{R}$ is the Gaussian Gram matrix of $q$. The ``mean-squared error'' is $\mathbf{P}^\intercal \mathbf{R}^{-1}\mathbf{P}$, a scalar to be optimized.
\item Now also view the data density $p$ as Gaussian residuals $p_1, p_2,\cdots,p_N$ for a batch of samples $X_1,X_2,\cdots,X_N$. Each residual is $p_n = \mathcal{N}(X-X_n;v_X)$ defined on one data sample. The error of using the two series of residuals, $q_1,q_2,\cdots,q_K$ and $p_1,p_2,\cdots,p_K$, for predicting each other is given by the trace of the Schur complement $Trace(\mathbf{R}_G^{-\frac{1}{2}}\mathbf{P}_{FG}^\intercal \mathbf{R}_F^{-1} \mathbf{P}_{FG} \mathbf{R}_G^{-\frac{1}{2}})$, where $\mathbf{R}_G$ and $\mathbf{R}_F$ are two Gaussian Gram matrices for $p$ and $q$.
\item We can also directly perform SVD on Gaussian cross Gram matrix $\mathbf{P}_{FG}$ and maximize the sum of its singular values, which has the best results. It turns out that this sum, the nuclear norm of $\mathbf{P}_{FG}$, can be viewed as a form of divergence.
\end{enumerate}

\noindent We name these costs based on Gaussian Gram matrices and cross Gram matrices the family of kernelized matrix costs, including the scalar cost $\frac{\langle p, q \rangle^2}{\langle q ,q\rangle}$, the vector-matrix cost $\mathbf{P}^\intercal \mathbf{R}^{-1}\mathbf{P}$, the trace of Schur complement $Trace(\mathbf{R}_G^{-\frac{1}{2}}\mathbf{P}_{FG}^\intercal \mathbf{R}_F^{-1} \mathbf{P}_{FG} \mathbf{R}_G^{-\frac{1}{2}})$, and the nuclear norm (the sum of singular values) of $\mathbf{P}_{FG}$. The nuclear norm cost offers the best performance.

Gaussian Gram matrix-based statistical measures can be traced back to KICA~\cite{bach2002kernel, giraldo2014measures, giraldo2012reproducing}, HSIC~\cite{gretton2005measuring}, and DCCA~\cite{andrew2013deep}. We propose costs for learning mixture densities using neural networks. The code for this paper is available at \url{https://github.com/bohu615/kernelized-matrix-cost}.
\vspace{-5pt}

\section{Essential Properties}\vspace{-5pt}

% Our proposal is made possible by the properties of Gaussian mixtures that the norm of a Gaussian mixture density has a closed form determined only by the mean, variance, and weights; the inner product of two Gaussian mixture densities also has a closed form.

Our proposal is made possible by the properties of Gaussian mixtures that the norm of a Gaussian mixture density has a closed form determined only by the mean, variance, and weights; the inner product of two Gaussian mixture densities also has a closed form, presented as follows: Property~\ref{property_1} shows the inner product between two Gaussian functions; Property~\ref{property_2} shows closed forms for norms and inner products of Gaussian mixtures with discrete priors; Property~\ref{property_3} shows closed forms for continuous priors.\vspace{-3pt}

\begin{property}
Given two Gaussian density functions $p_1(X) = \mathcal{N}(X-m_1;v_1)$ and $p_2(X) = \mathcal{N}(X-m_2;v_2)$, their inner product has a closed form:\vspace{-10pt}
\begin{equation}
\resizebox{.5\linewidth}{!}{
$\begin{gathered}
\langle p_1, p_2 \rangle = \mathcal{N}(m_1-m_2; {v_1+v_2}).
\end{gathered}$}
\end{equation}
\label{property_1}
\end{property}\vspace{-20pt}

\begin{property}
(Gaussian mixtures with discrete priors.) Given a discrete Gaussian mixture $p(X) = \sum_{k=1}^K w_k \mathcal{N}(X - m_k;v_k)$, the $L_2$ norm of $p$ satisfies\vspace{-10pt}
\begin{equation}
\resizebox{.6\linewidth}{!}{
$\begin{gathered}
\langle p ,p\rangle = \sum_{i=1}^K \sum_{j=1}^K w_i w_j\mathcal{N}(m_i-m_j;v_i+v_j).\vspace{-5pt}
\end{gathered}$}
\end{equation}
Given another mixture $q(X) = \sum_{k=1}^{K'} w_k' \mathcal{N}(X - m_k';v_k')$, the inner product between $p$ and $q$ satisfies
\begin{equation}
\resizebox{.6\linewidth}{!}{
$\begin{gathered}
\langle p ,q\rangle = \sum_{i=1}^K \sum_{j=1}^{K'} w_i w_j'\mathcal{N}(m_i-m_j';v_i+v_j').
\end{gathered}$}
\end{equation}\vspace{-10pt}
\label{property_2}
\end{property}

\begin{property}
(Gaussian mixtures with any priors.) Given a Gaussian mixture with a prior distribution $p(X) = \int p(c) w(c) \mathcal{N}(X - m(c);v(c)) dc$. The norm of $p(X)$ satisfies
\begin{equation}
\resizebox{.9\linewidth}{!}{
$\begin{aligned}
& \langle p, p\rangle = \iint p(c_1)p(c_2) w(c_1)w(c_2) \\ & \;\;\;\;\;\;\;\;\;\;\;\;\;\;\;\;\;\;\;\;\mathcal{N}(m(c_1)-m(c_2);v(c_1)+v(c_2)) dc_1dc_2\\
& \;\;\;\;\;\;\;\;\;= \mathbb{E}_{\mathbf{c}_1,\mathbf{c}_2}\left[ w(\mathbf{c}_1)w(\mathbf{c}_2)  \mathcal{N}(m(\mathbf{c}_1)-m(\mathbf{c}_2); v(\mathbf{c}_1)+v(\mathbf{c}_2))\right].\vspace{15pt}
\end{aligned}$}
\end{equation}
\noindent Given another Gaussian mixture $q(X) = \int p'(c) w'(c)\mathcal{N}(X-m'(c);v'(c)) dc$, the inner product between them satisfies\vspace{-5pt}
\begin{equation}
\resizebox{.9\linewidth}{!}{
$\begin{aligned}
& \langle p, q\rangle = \iint p(c)p'(c') w(c)w'(c') \\ & \;\;\;\;\;\;\;\;\;\;\;\;\;\;\;\;\;\;\;\;\mathcal{N}(m(c)-m'(c');v(c)+v'(c')) dc dc'\\
&\;\;\;\;\;\;\;\;\; = \mathbb{E}_{\mathbf{c},\mathbf{c}'}\left[ w(\mathbf{c})w'(\mathbf{c}')  \mathcal{N}(m(\mathbf{c})-m'(\mathbf{c}'); v(\mathbf{c})+v'(\mathbf{c}'))\right]. 
\end{aligned}$}
\end{equation}\vspace{-10pt}
\label{property_3}
\end{property}

\begin{corollary}
The $L_p$ norm of a Gaussian mixture for any exponent, regardless of discrete or continuous prior, has a closed form.
\end{corollary}\vspace{-4pt}

Thus, norms and inner products of Gaussian mixtures, with discrete priors $p(X) = \sum_{k=1}^K w_k \mathcal{N}(X - m_k;v_k)$ or arbitrary priors $p(X) = \int p(c) w(c) \mathcal{N}(X - m(c);v(c)) dc$, have closed forms determined by mean distances, variance sums, and weight products, which is a double sum for discrete cases and an expectation for continuous cases.

\section{Kernelized Matrix Costs}

With the closed-form solutions for Gaussian functions and mixture densities, we propose the following costs for using a Gaussian mixture $q$ to approximate a data density $p$:
\begin{enumerate}[leftmargin=*]
\item Proposition~\ref{proposition4}, $sc(q;p)$: Apply the Schwarz inequality directly;
\item Proposition~\ref{proposition5}, $vc(\mathbf{f};p)$: Suppose we have a series of Gaussian residuals $\mathbf{f} = [q_1, q_2,\cdots,q_K]^\intercal$. What is the optimal linear weights and ``mean-squared error'' for them to predict a density $p$?
\item Proposition~\ref{proposition6}, $mc(\mathbf{f}, \mathbf{g};p)$: Suppose we have two series of residuals, $\mathbf{f} = [q_1, q_2,\cdots,q_K]^\intercal$ for the model and $\mathbf{g} = [p_1,p_2,\cdots,p_N]^\intercal$ for the data. What is the error for them to predict each other?
\item Proposition~\ref{proposition7}: Perform the SVD on the Gaussian cross Gram matrix $\mathbf{P}_{FG}$ directly and maximize its singular values.\vspace{-3pt}
\end{enumerate}

\begin{prop}
(Scalar Cost.) Given a model density $q$ and a data density $p$. By the Schwarz inequality,\vspace{-5pt}
\begin{equation}
\begin{gathered}
\langle p, q \rangle^2 \leq \langle p,p \rangle \cdot \langle q,q \rangle.\vspace{-4pt}
\end{gathered}
\end{equation}
\noindent Define the cost $sc(q;p)$ as follows with an upper bound\vspace{-5pt}
\begin{equation}
\begin{gathered}
 sc(q;p) = \frac{\langle p, q \rangle^2}{\langle q, q \rangle},\; sc(q;p) \leq \langle p, p \rangle.\vspace{-4pt}
\end{gathered}
\end{equation}
The upper bound is tight when $q=p$.\vspace{-3pt}
\label{proposition4}
\end{prop}

\begin{prop}
(Vector-Matrix Cost.) Given a series of Gaussian residuals $\mathbf{f} = [q_1, q_2,\cdots,q_K]^\intercal$. Build an auto-correlation matrix $\mathbf{R} = \int \mathbf{f}\mathbf{f}^\intercal dX$ and a cross-correlation vector $\mathbf{P} = \int \mathbf{f} \cdot p\;  dX$. We define the vector-matrix cost as\vspace{-3pt}
\begin{equation}
\begin{gathered}
vc(\mathbf{f}; p) = \mathbf{P}^\intercal \mathbf{R}^{-1} \mathbf{P},\;  vc(\mathbf{f}; p) \leq \langle p, p \rangle.\vspace{-3pt}
\end{gathered}
\end{equation}\vspace{-15pt}
\label{proposition5}
\end{prop}

\begin{prop}
(Matrix-Matrix Cost.) Given two series of Gaussian residuals $\mathbf{f} = [q_1,q_2,\cdots,q_K]^\intercal$ and $\mathbf{g} = [p_1,p_2,\cdots,p_N]^\intercal$. Build two auto-correlation matrices and their cross-correlation matrix using:\vspace{-3pt}
\begin{equation}
\resizebox{.6\linewidth}{!}{
$\begin{gathered}
\mathbf{R}_F = \int \mathbf{f}\mathbf{f}^\intercal dX,\; \mathbf{R}_G = \int \mathbf{g}\mathbf{g}^\intercal dX,\\ \mathbf{P}_{FG} = \int \mathbf{f} \mathbf{g}^\intercal dX, \; \mathbf{R}_{FG} = \begin{bmatrix}\mathbf{R}_F & \mathbf{P}_{FG} \\
\mathbf{P}^\intercal_{FG} & \mathbf{R}_G
\end{bmatrix}.
\end{gathered}$}
\end{equation}
We maximize the trace of the Schur complement\vspace{-3pt}
\begin{equation}
\resizebox{.77\linewidth}{!}{
$\begin{gathered}
mc(\mathbf{f}, \mathbf{g};p) = Trace(\mathbf{R}_G^{-\frac{1}{2}}\mathbf{P}_{FG}^\intercal \mathbf{R}_F^{-1} \mathbf{P}_{FG} \mathbf{R}_G^{-\frac{1}{2}}). 
\end{gathered}$}
\end{equation}
or equivalently minimizing the log-determinant $\log\det \mathbf{R}_{FG} - \log\det \mathbf{R}_{F} - \log\det \mathbf{R}_{G}.$ The two are interchangeable and we use the trace cost for analysis. Suppose the data are fixed and we train only the model, the matrix $\mathbf{R}_G$ can be ignored.
\label{proposition6}
\end{prop}

\begin{prop}
(SVD Cost.) For a batch of data samples $X_1, X_2, \cdots, X_N$ and a batch of centers $X_1', X_2', \cdots, X'_K$ produced by a neural net, we directly build a Gaussian cross-correlation matrix $\mathbf{P}_{FG}$ between them and perform SVD. The objective is to maximize its singular values:\vspace{-7pt}
\begin{equation}
\resizebox{.8\linewidth}{!}{
$\begin{gathered}
\mathbf{P}_{FG}  = \mathbf{U}\mathbf{S}\mathbf{V},\;
\mathbf{U}\mathbf{U}^\intercal = \mathbf{I}, \; \mathbf{V}\mathbf{V}^\intercal = \mathbf{I}, \; \mathbf{S} = \begin{bmatrix}
    \sigma_{1} &  \\
    & \ddots  \\
    & & \sigma_{N}
\end{bmatrix}, \vspace{-13pt} \\ 
\text{max} \;\; \sum_{k=1}^K \sigma_k.\vspace{-3pt}
\end{gathered}$}
\label{singular_values}
\end{equation}
We found that decomposing a Gaussian cross Gram matrix $\mathbf{P}_{FG}$ with a small variance is a must. Decomposing an $L_2$ distance matrix is ineffective, as is the Frobenius norm $Trace(\mathbf{P}_{FG}\mathbf{P}_{FG}^\intercal)$.\vspace{-3pt}
\label{proposition7}
\end{prop}

We further explain how to apply them. For all costs, a series of centers is required to define the mixture. At each training step, sample noise $u_1,\cdots,u_K$ from a prior distribution (uniform, Gaussian, or hybrid). Next, a neural network maps the noise to generated samples $X_1',X_2',\cdots,X_K'$. Sample a batch of samples $X_1,X_2,\cdots,X_N$ at each iteration. Approximate the data and model densities with\vspace{-4pt}
\begin{equation}
\resizebox{.9\linewidth}{!}{
$\begin{gathered}
p(X) \approx \frac{1}{N}\sum_{n=1}^N \mathcal{N}(X-X_n;v_p), \; q(X) \approx \frac{1}{K}\sum_{k=1}^K \mathcal{N}(X-X_k';v_q).
\end{gathered}$}
\label{pair_wise_difference}
\end{equation}
Then for the scalar cost, the norms and the inner product follow\vspace{-5pt}
\begin{equation}
\resizebox{1\linewidth}{!}{
$\begin{gathered}
\langle p, q\rangle = \frac{1}{NK}\sum_{n=1}^N \sum_{k=1}^K \mathcal{N}(X_n - X_k';v_p+v_q), \\
\langle q, q\rangle = \frac{1}{K^2} \sum_{i=1}^K \sum_{j=1}^K \mathcal{N}(X_i' - X_j';2v_q), \;\langle p, p\rangle = \frac{1}{N^2} \sum_{i=1}^N \sum_{j=1}^N \mathcal{N}(X_i - X_j;2v_p).
\end{gathered}$}
\end{equation}

\noindent For the matrix costs, a Gaussian cross-correlation matrix can be constructed by\vspace{-1pt}
\begin{equation}
\resizebox{1\linewidth}{!}{
$\begin{gathered}
\mathbf{M}_{FG} =
\frac{1}{d_X}\begin{bmatrix}
||X_1-X_1'||_2^2 & \cdots & ||X_1-X_K'||_2^2 \\
\vdots      & \ddots & \vdots  \\
||X_N-X_1'||_2^2 & \cdots & ||X_N-X_K'||_2^2  
\end{bmatrix}, \;\mathbf{P}_{FG} \approx \exp(-\frac{1}{2(v_p+v_q)}\mathbf{M}_{FG}).
\end{gathered}$}
\label{gram_matrices}
\end{equation}
That is, we first construct the matrix of $L_2$ distances $\mathbf{M}_{FG}$ between all pairs of $X_n$ and $X_k'$, divide it by data dimension $d_X$, scale by the sum of variances $v_p+v_q$, and take its exponential. For simplicity, one can set $v_p=v_q=v$. Due to normalization, the Gaussian pdf's scalar constant can be ignored, as it is also arbitrarily small in high dimensions. Scaling with $d_X$ is crucial for numerical stability in high dimensions. The Gaussian auto-correlation $\mathbf{R}_F$ and $\mathbf{R}_G$ can be constructed similarly. For the vector cost, the expectation $\mathbf{P} = \int \mathbf{f} \cdot p\;  dX$ can be obtained simply by summing rows of $\mathbf{P}_{FG}$.

With approximated norms, Gram matrices, and expectations, we can build the costs from propositions. We found that the scalar cost, vector-matrix cost, and matrix-matrix cost perform similarly, while the SVD cost outperforms the others.

% Scalar cost, vector-matrix cost, and matrix-matrix cost perform similarly, whereas the SVD cost outperforms the others.

% Having approximated the norms, Gram matrices $\mathbf{R}_F$, $\mathbf{R}_G$, $\mathbf{P}_{FG}$ and the expectation $\mathbf{P}$, we can build the costs introduced in the propositions. We found that the scalar cost, the vector-matrix cost, and the matrix-matrix cost performs very similarly, while the SVD cost has better performance than the others. 

\section{Theoretical Justification}

Because of the Schwarz inequality, the scalar cost $sc(q;p)$ is naturally upper bounded by the norm of $p$. Because both the vector-matrix cost $vc(\mathbf{f}; p) = \mathbf{P}^\intercal \mathbf{R}^{-1} \mathbf{P}$ and matrix-matrix cost $mc(\mathbf{f}, \mathbf{g};p) = Trace(\mathbf{R}_G^{-\frac{1}{2}}\mathbf{P}_{FG}^\intercal \mathbf{R}_F^{-1} \mathbf{P}_{FG} \mathbf{R}_G^{-\frac{1}{2}})$ are defined with the optimal linear predictor of $p$, it can be shown that they are also upper bounded by the norm of $p$.

Note that $\mathbf{P} = \int \mathbf{f}\cdot p ;dX$ in in the vector-matrix cost is an expectation vector of Gaussian residuals, and $\mathbf{P}_{FG} = \int  \mathbf{f}  \mathbf{g}^\intercal dX$ in the matrix-matrix cost is a Gaussian cross-correlation matrix by treating both the data and the model densities as Gaussian residuals. 

Using $vc = \mathbf{P}^\intercal \mathbf{R}^{-1} \mathbf{P}$ as an example. When it is maximized, the prediction of the density $p$ with the series of Gaussian residuals $\mathbf{f}$ is $q = (\mathbf{R}^{-1}\mathbf{P})^\intercal \mathbf{f}\approx p$, Then the cost $vc$ becomes $vc = \int (\mathbf{R}^{-1}\mathbf{P})^\intercal \mathbf{f} \cdot p\;dX = \int q\cdot p\; dX \approx \int p^2 dX $, which is also upper bounded by the norm of $p$. The same analysis can be applied to the matrix-matrix cost $mc(\mathbf{f}, \mathbf{g};p) = Trace(\mathbf{R}_G^{-\frac{1}{2}}\mathbf{P}_{FG}^\intercal \mathbf{R}_F^{-1} \mathbf{P}_{FG} \mathbf{R}_G^{-\frac{1}{2}})$. 

A simplified justification for maximizing singular values of $\mathbf{P}_{FG}$ is that when samples match one-by-one, with $X_n=X_n'$, the cross-correlation $\mathbf{P}_{FG}$ will become an auto-correlation matrix, thus Hermitian. In this case, the sum of its singular values is the sum of its eigenvalues, which is also the matrix trace. The diagonal elements of a Hermitian Gaussian Gram matrix are all constants $\mathcal{N}(X_n-X_n) = \mathcal{N}(0)$, so the trace is $N\cdot \mathcal{N}(0)$. We found that this trace, a constant $N\cdot \mathcal{N}(0)$, is the maximal value that the cost can reach. Though the solution $X_n =X_n'$ is non-unique, when $p(X)$ and $q(X)$ are far apart, the nuclear norm of $\mathbf{P}_{FG}$ will be smaller than this constant value. The detailed analysis is as follows.\vspace{-5pt}

\begin{property}
With a continuous kernel function $\mathcal{K}(X,X')$ and density functions $p(X)$, $q(X)$, we propose two decompositions based on Mercer's theorem: decomposing $\sqrt{p(X)}\mathcal{K}(X,X')\sqrt{q(X')}$ with orthonormal bases w.r.t. Lebesgue measure $\mu$ (Eq.~\eqref{decomposition1}); and decomposing $\mathcal{K}(X,X')$ with bases orthonormal w.r.t. probability measures $p$, $q$ (Eq.~\eqref{decomposition2}). The two decompositions share the same singular values. Their discrete equivalents for Hermitian matrices $\mathbf{K}_{XX'}$ and discrete densities are shown in Eq.~\eqref{discrete1} and Eq.~\eqref{discrete2}.\vspace{-7pt}
\begin{equation}
\resizebox{.65\linewidth}{!}{
$\begin{gathered}
    \sqrt{p(X)}\mathcal{K}(X,X')\sqrt{q(X')} = \sum_{k=1}^K \lambda_k {\phi_k}(X){\psi_k}(X'),\vspace{-3pt}\\ \vspace{-3pt}
\int {\phi_i}{\phi_j} dX = \int {\psi_i}{\psi_j} dX' = \begin{cases} 1, \;i = j& \\ 0, \; i\neq j \end{cases}\hspace{-9pt}. 
\end{gathered}$}
\label{decomposition1}
\end{equation}\vspace{-5pt}
\begin{equation}
\resizebox{.65\linewidth}{!}{
$\begin{gathered}
\mathcal{K}(X,X') = \sum_{k=1}^K \lambda_k \widehat{\phi_k}(X) \widehat{\psi_k}(X'),\\
\int \widehat{\phi_i} \widehat{\phi_j} p(X) dX = \int \widehat{\psi_i} \widehat{\psi_j} q(X') dX' = \begin{cases} 1, \;i = j& \\ 0, \; i\neq j \end{cases}\hspace{-7pt}. 
\end{gathered}$}
\label{decomposition2}
\end{equation}
\begin{equation}
\resizebox{.9\linewidth}{!}{
$\begin{gathered}
{diag}(\sqrt{\mathbf{P}_X}) \, \mathbf{K}_{X,X'} \, diag(\sqrt{\mathbf{Q}_{X
'}}) = \mathbf{U}\mathbf{S}\mathbf{V},\;\;
\mathbf{U}\mathbf{U}^\intercal  = \mathbf{I}, \;\; \mathbf{V}\mathbf{V}^\intercal = \mathbf{I}. 
\end{gathered}$}
\label{discrete1}
\end{equation}\vspace{-10pt}
\begin{equation}
\resizebox{.9\linewidth}{!}{
$\begin{gathered}
\mathbf{K}_{X,X'} = \mathbf{U}\mathbf{S}\mathbf{V}, \, \mathbf{U} diag(\mathbf{P}_X)\, \mathbf{U}^\intercal  = \mathbf{I}, \, \mathbf{V} diag(\mathbf{Q}_X) \mathbf{V}^\intercal = \mathbf{I}. 
\end{gathered}$}
\label{discrete2}
\end{equation}
\noindent Since the matrix $\mathbf{K}_{XX'}$ is Hermitian, we can decompose it with $\mathbf{K}_{XX'} = \mathbf{Q}_{\mathbf{N}}\mathbf{\Lambda}_{\mathbf{N}}\mathbf{Q}_{\mathbf{N}}$. Define $\mathbf{A}:= diag(\sqrt{\mathbf{P}_X}) \mathbf{Q}_{\mathbf{N}} \mathbf{\Lambda}_{\mathbf{N}}^{\frac{1}{2}}$ and $\mathbf{B} :=  diag(\sqrt{\mathbf{Q}_{X}}) \mathbf{Q}_{\mathbf{N}} \mathbf{\Lambda}_{\mathbf{N}}^{\frac{1}{2}}$, applying the inequality of the nuclear norm:\vspace{-2pt}
\begin{equation}
\resizebox{.7\linewidth}{!}{
$\begin{aligned}
& \hspace{-6pt} ||\mathbf{A}\mathbf{B}^\intercal||_*  \leq \sqrt{||\mathbf{A}\mathbf{A}^\intercal||_*} \cdot \sqrt{||\mathbf{B}\mathbf{B}^\intercal||_*}, \\
||\mathbf{A}\mathbf{A}^\intercal||_* &=  ||{diag}(\sqrt{\mathbf{P}_X}) \mathbf{K}_{XX'} diag(\sqrt{\mathbf{P}_{X}})||_* \\
&= Trace({diag}(\sqrt{\mathbf{P}_X}) \mathbf{K}_{XX'} diag(\sqrt{\mathbf{P}_{X}})) \\
&= \mathcal{N}(0) = ||\mathbf{B}\mathbf{B}^\intercal||_*. 
\end{aligned}$}\vspace{-3pt}
\label{bound_AB}
\end{equation}
\noindent That is, the nuclear norm of the defined matrix ${diag}(\sqrt{\mathbf{P}_X}) \mathbf{K}_{XX'} diag(\sqrt{\mathbf{Q}_{X}})$ is upper bounded by the constant $\mathcal{N}(0)$. The bound is tight when $\mathbf{A} = \mathbf{B}$  for positive eigenvalues of $\mathbf{K}_{XX'}$, i.e., when $diag(\sqrt{\mathbf{P}_X}) \mathbf{Q}_{\mathbf{N}} =  diag(\sqrt{\mathbf{Q}_{X}}) \mathbf{Q}_{\mathbf{N}}$.
\label{property8}
\end{property}\vspace{-3pt}

Property~\ref{property8} shows the decomposition of a continuous function $\sqrt{p(X)}\mathcal{K}(X,X')\sqrt{q(X')}$ using Mercer's theorem (Eq.~\eqref{decomposition1}). Suppose $\phi_i$ and $\psi_i$ are the bases of this function, by applying variational trick $\widehat{\phi_i} = \phi_i / \sqrt{p(X)}$ and $\widehat{\psi_i} = \psi_i / \sqrt{q(X)}$, we obtain bases $\widehat{\phi_i}$ and $\widehat{\psi_i}$ orthonormal to the probability measures $p(X)$ and $q(X)$ that decompose the kernel $\mathcal{K}(X,X')$ (Eq.~\eqref{decomposition2}). In summary, the variational trick transforms the decomposition of $\sqrt{p(X)}\mathcal{K}(X,X')\sqrt{q(X')}$ (Eq.~\eqref{decomposition1}) into decomposing $\mathcal{K}(X,X')$ (Eq.~\eqref{decomposition2}), by changing the measures from the Lebesgue measure to probability measures. This decomposition of $\mathcal{K}(X,X')$ with bases orthonormal to probability measures is the SVD of the Gaussian cross-correlation matrix $\mathbf{P}_{FG}$.  

% This decomposition of $\mathcal{K}(X,X')$ with bases orthonormal to probability measures is the decomposition of the Gaussian cross-correlation matrix $\mathbf{P}_{FG}$ in the SVD cost.  

% Let $\mathbf{P}_{FG}(m,n)$ denote matrix elements, with singular vectors $U(m)$, $V(n)$. The SVD, $\mathbf{P}_{FG}(m,n) = \mathcal{N}(X_m-X_n) = \sum_{i=1}^N \sigma_i U_i(m)V_i(n)$, is decomposing $\mathcal{K}(X,X')$ as a Gaussian kernel.

% The decomposition of $\mathcal{K}(X,X')$ with bases orthonormal to the probability measures, after the variational trick is applied, exactly matches the decomposition of the Gaussian cross-correlation matrix $\mathbf{P}_{FG}$ in the SVD cost. Let $\mathbf{P}_{FG}(m,n)$ denote matrix elements, with singular vectors $U(m)$, $V(n)$. The SVD, $\mathbf{P}_{FG}(m,n) = \mathcal{N}(X_m-X_n) = \sum_{i=1}^N \sigma_i U_i(m)V_i(n)$, exactly corresponds to decomposing $\mathcal{K}(X,X')$ as a Gaussian kernel.

The inspiration also comes from the following. A conventional $f$-divergence~\cite{jordan1, jordan2} is a functional on the density ratio $\frac{p(X)}{q(X)}$. Suppose the densities are discrete. This ratio is a vector that does not have a standard convenient orthonormal decomposition. But if we define a quantity like ${diag}(\sqrt{\mathbf{P}_X}) \mathbf{K}_{XX'} diag(\sqrt{\mathbf{Q}_{X}})$, then the decomposition becomes possible. If we pick $\mathbf{K}_{X,X'}$ to be an identity matrix, correspondingly the identify function $\mathcal{K}(X,X') = \mathbbm{1}\{X=X'\}$, then the matrix to decompose becomes ${diag}(\sqrt{\mathbf{P}_X}) \;diag(\sqrt{\mathbf{Q}_{X}})$, a diagonal matrix with elements $\sqrt{P_X} \sqrt{Q_X}$. Summing its singular values becomes $\int \sqrt{p(X)}\cdot \sqrt{q(X)}\; dX$, a form of Hellinger distance. One can spot the drawback that for continuous $p$ and $q$, this decomposition will generate an infinite number of singular values at each point in the sample domain when $\sqrt{p(X)}\sqrt{q(X)}$ is positive. So a smoother like a Gaussian function as $\mathcal{K}(X,X')$ is a must such that we can still measure the distance between $\sqrt{p(X)}$ and $\sqrt{q(X)}$ but with a finite number of singular values.

It is also possible to discuss the optimal singular functions when $q = p$. Not only can we can apply eigendecomposition $\mathbf{K}_{XX'} = \mathbf{Q}_{\mathbf{N}}\mathbf{\Lambda}_{\mathbf{N}}\mathbf{Q}_{\mathbf{N}}$, but we can also decompose a Gaussian function using the closed-form inner product $\mathcal{N}(X-X';2v) = \int \mathcal{N}(X-X'';v)\mathcal{N}(X'-X'';v) dX''$. So $\mathbf{K}_{XX'}$ can also be decomposed as $\mathbf{K}_{XX'} = \mathbf{K}_{XX''}\mathbf{K}_{XX''}^\intercal$, with $\mathbf{K}_{XX''}$ having half of the variance of $\mathbf{K}_{XX'}$. Denote $\mathbf{C} = diag(\sqrt{\mathbf{P}_X}) \mathbf{K}_{XX''}$, then\vspace{-3pt}
\begin{equation}
\begin{aligned}
{diag}(\sqrt{\mathbf{P}_X}) \mathbf{K}_{XX'} diag(\sqrt{\mathbf{P}_{X}}) = \mathbf{C}\mathbf{C}^\intercal,
\end{aligned}\vspace{-3pt}
\label{CCT}
\end{equation}
implying that the eigenvector of ${diag}(\sqrt{\mathbf{P}_X}) \mathbf{K}_{XX'} diag(\sqrt{\mathbf{P}_{X}})$ must match the left singular vector of $\mathbf{C}$. One can further see that $\mathbf{C}$ represents $\sqrt{p(X)}\mathcal{N}(X-X'';v)$, the square root of a joint density $p(X,X'') = p(X) \mathcal{N}(X-X'';2v)$ up to a constant, representing the data density $p(X)$ passed through a Gaussian conditional.

Additionally, the inner product between the model residuals $\mathbf{f} = [q_1,q_2,\cdots,q_K]^\intercal$ and the data residuals $\mathbf{g} = [f_1,f_2,\cdots,f_N]^\intercal$ is given by $\mathbf{P}_{FG} = \int \mathbf{f}\mathbf{g}^\intercal \;dX$. Given SVD $\mathbf{P}_{FG} = \mathbf{U}\mathbf{S}\mathbf{V}$, we can further normalize $\widehat{\mathbf{f}} = \mathbf{U}^\intercal \mathbf{f}$ and $\widehat{g} = \mathbf{V}^\intercal \mathbf{g}$. If we write $\mathbf{P}_{FG}$ as \vspace{-3pt}
\begin{equation}
\resizebox{0.8\linewidth}{!}{
$\begin{gathered}
\mathbf{P}_{FG} = \int \mathbf{f}\mathbf{g}^\intercal \;dX = \iint \mathbf{f}(X')\mathbbm{1}\{X'=X\} \mathbf{g}^\intercal (X) dX'dX.\\
\iint \widehat{\mathbf{f}}(X') \mathbbm{1}\{X'=X\} \widehat{\mathbf{g}}^\intercal (X) dX' dX = \mathbf{S},\vspace{-5pt}
\end{gathered}$}
\end{equation}
meaning that we put an identity function $\mathbbm{1}\{X'=X\}$ between $\mathbf{f}$ and $\mathbf{g}$ to make the integral over $X$ a double integral. Then, the normalized functions $\widehat{f}(X')$ and $\widehat{g}(X)$ can be said to decompose $\mathbbm{1}\{X'=X\}$:\vspace{-7pt}
\begin{equation}
\resizebox{0.5\linewidth}{!}{
$\begin{gathered}
\mathbbm{1}(X'=X) \approx \sum_{k=1}^K \sigma_k \widehat{\mathbf{f}}_k(X') \widehat{\mathbf{g}}_k(X),\vspace{-4pt}
\end{gathered}$}
\label{quantity}
\end{equation}
which can be described as using $\widehat{f}$ and $\widehat{\mathbf{g}}$, the Gaussian residuals with affine transformations $\mathbf{U}$ and $\mathbf{V}$, to approximate and decompose the identity function $\mathbbm{1}\{X=X'\}$, i.e., using Gaussian residuals to come up with the best approximator of the identity $\mathbbm{1}\{X=X'\}$. This conclusion of approximating identity function with affine transformed Gaussians also applies to the vector-matrix and matrix-matrix costs.\vspace{-7pt}
\begin{property}
If we maximize $Trace(\mathbf{R}_G^{-\frac{1}{2}}\mathbf{P}_{FG}^\intercal \mathbf{R}_F^{-1} \mathbf{P}_{FG} \mathbf{R}_G^{-\frac{1}{2}})$, apply the eigendecomposition and transformations\vspace{-7pt}
\begin{equation}
\resizebox{.85\linewidth}{!}{
$\begin{aligned}
    \mathbf{R}_F^{-\frac{1}{2}} \mathbf{P}_{FG} \mathbf{R}_G^{-\frac{1}{2}} = \mathbf{U}\mathbf{S}\mathbf{V}, \;\widehat{\mathbf{f}}^* = \mathbf{U}^\intercal\mathbf{R}_F^{-\frac{1}{2}}\mathbf{f}^*, \;\widehat{\mathbf{g}} = \mathbf{V}^\intercal \mathbf{R}_G^{-\frac{1}{2}} \mathbf{g}.
\end{aligned}$}\vspace{-5pt}
\end{equation}
In this case, it can be shown that $\widehat{\mathbf{f}}^*$ and $\mathbf{\widehat{g}}$ is the best possible solution for approximating the identity $\mathbbm{1}\{X=X'\}$ using these residuals, with the added property of orthonormality for using $\mathbf{R}_F^{-\frac{1}{2}}, \mathbf{R}_G^{-\frac{1}{2}}$.
% It can also be shown that in this case, $\widehat{\mathbf{f}}^*$ and $\mathbf{\widehat{g}}$ are the best possible solution for approximating the identity function $\mathbbm{1}\{X=X'\}$ that can be found by these residuals. The difference is that now they are orthonormal functions.
\end{property}\vspace{-6pt}

\section{Multivariate Function Approximator}\label{section_multivariate}

We have introduced using the decomposition to approximate a kernel function $\mathcal{K}(X,X')$ of two arguments. In the high level, we are using a function approximator of a form $k(X,X') = \sum_{k=1}^K {\phi_k}(X){\psi_k}(X')$  (Eq.~\eqref{change_mixture_decoder}). Here we also propose a multivariate extension to $k$.\vspace{-4pt}

\begin{property}

For variables $X_1, X_2, \dots, X_T$, we initialize multivariate functions $\phi_k^{(1)}, \phi_k^{(2)}, \dots, \phi_k^{(T)}$ with $K$ entries each. We define a multivariate function $k(X_1, X_2, \dots, X_T)$ (Eq.~\eqref{change_mixture_decoder}) as the product of functions over $t$ and their sum over $k$. But a product of functions may be numerically unstable, so we make variational changes to define $\widehat{k}$:\vspace{-3pt}

\begin{itemize}[leftmargin=*]
\item Modify functions to $\phi_k(X_t, t)$ with sample and position as inputs;
\item Replace product with the exponential of the mean;
\item Use negative mean of the square to bound exponential value by $1$. Adding a real coefficient $\alpha$, since each Gaussian is bounded by $1$.\vspace{-15pt}
\end{itemize}
\end{property}

% % Given a series of variables $X_1,X_2,\cdots,X_T$, we initiate a series of multivariate functions $\phi_k^{(1)}, \phi_k^{(2)},\cdots,\phi_k^{(T)}$, each one of them having $K$ entries. We define in Eq.~\eqref{change_mixture_decoder} a multivariate function $k(X_1,X_2,\cdots,X_T)$ as the products over $t$ and the sum over $k$. 

% % One can spot that the product of a series of functions may not be numerical stable and the value of it may vanish. Thus make a few variational changes and define $\widehat{k}$ in Eq.~\eqref{change_mixture_decoder}:\vspace{-3pt}
% \begin{itemize}[leftmargin=*]
%     \item Simplify each function to $\phi_k(X_t, t)$, a universal approximator receiving a sample and a position;
%     \item Use the exponential of the mean, instead of the product;
%     \item Use the negative mean of the square, such that the value after the exponential is bounded by $1$. Since each Gaussian component is bounded by $1$, we also add an arbitrary real coefficient $\alpha$.\vspace{-15pt}
%     \end{itemize}
% \end{property}
\begin{equation}
\resizebox{.7\linewidth}{!}{
$\begin{gathered}
k(X,X') = \sum_{k=1}^K {\phi_k}(X){\psi_k}(X').\\
\Downarrow \vspace{-6pt} \\ \vspace{-6pt}
k(X_1,X_2,\cdots,X_T) = \sum_{k=1}^K \phi_k^{(1)}(X_1) \phi_k^{(2)}(X_2) \cdots \phi_k^{(T)}(X_T).\\ \Downarrow \vspace{-3pt} \\ \vspace{-6pt}
\widehat{k}(X_1,X_2,\cdots,X_T) = \alpha \cdot \sum_{k=1}^K \exp\left(-\frac{1}{T}\sum_{t=1}^T\widehat{\phi_k}{}^2(X_t, t)\right).\vspace{-2pt}
\end{gathered}$}
\label{change_mixture_decoder}
\end{equation}
Unlike a standard network, this topology first maps pixels and patches to multivariate features that do not interact until the final exponential of averages. We still use batch normalization to improve results, but applied only to features of each pixel or patch without creating interactions. To achieve universality, a series of layers denoted as $\mathbf{k}^{(1)}, \mathbf{k}^{(2)}, \cdots, \mathbf{k}^{(M)}$ is applied to each pixel or patch $\mathbf{x}(t)$ (Eq.~\eqref{equation1}), which we also force to be Gaussian functions. Starting with $\mathbf{y}^{(0)}(t) = \mathbf{x}(t)$, layers are applied sequentially. Each layer $\mathbf{k}^{(m)}$ contains an affine transformation $\mathbf{A}^{(m)}$, weights $\mathbf{\mathcal{W}}^{(m)}(t)$ as Gaussian center anchors, an exponential of the $L_2$, and a batch norm (Eq.~\eqref{equation2}). The interaction among $t$ happens only in the final layer (Eq.~\eqref{equation3}).\vspace{-5pt}
\begin{equation}
\begin{aligned}
\mathbf{k}_t(\mathbf{x}(t)) = \mathbf{k}^{(M)}\cdots (\mathbf{k}^{(2)}(\mathbf{k}^{(1)}(\mathbf{x}(t)))).
\end{aligned}
\label{equation1}
\end{equation}
\begin{equation}
\begin{aligned}
\mathbf{k}^{(m)}(\mathbf{y}^{(m-1)}(t)) = BN\left(e^{-||\mathbf{A}^{(m)} \mathbf{y}^{(m-1)}(t) - \mathbf{\mathcal{W}}^{(m)}(t)||_2^2} \right).
\end{aligned}
\label{equation2}
\end{equation}\vspace{-10pt}
\begin{equation}
\begin{aligned}
k(\mathbf{x}_1, \mathbf{x}_2, \cdots \mathbf{x}_T) = \mathbf{A}_F e^{-\frac{1}{T}\sum_{t=1}^T || \mathbf{k}_t(\mathbf{x}(t)) - \mathbf{\mathcal{W}}_F(t)||_2^2}.
\end{aligned}
\label{equation3}
\end{equation}\vspace{-20pt}

\section{Experiments}\vspace{-1pt}

\noindent \textbf{\textit{Generations.}} We found no difficulties in using costs to train a standard mixture density network to fit toy 2D datasets and image datasets like MNIST and CelebA. Among the costs, the SVD cost works best.

\begin{figure}[h]
\centering
\begin{subfigure}{.24\textwidth}\includegraphics[width=.9\linewidth]{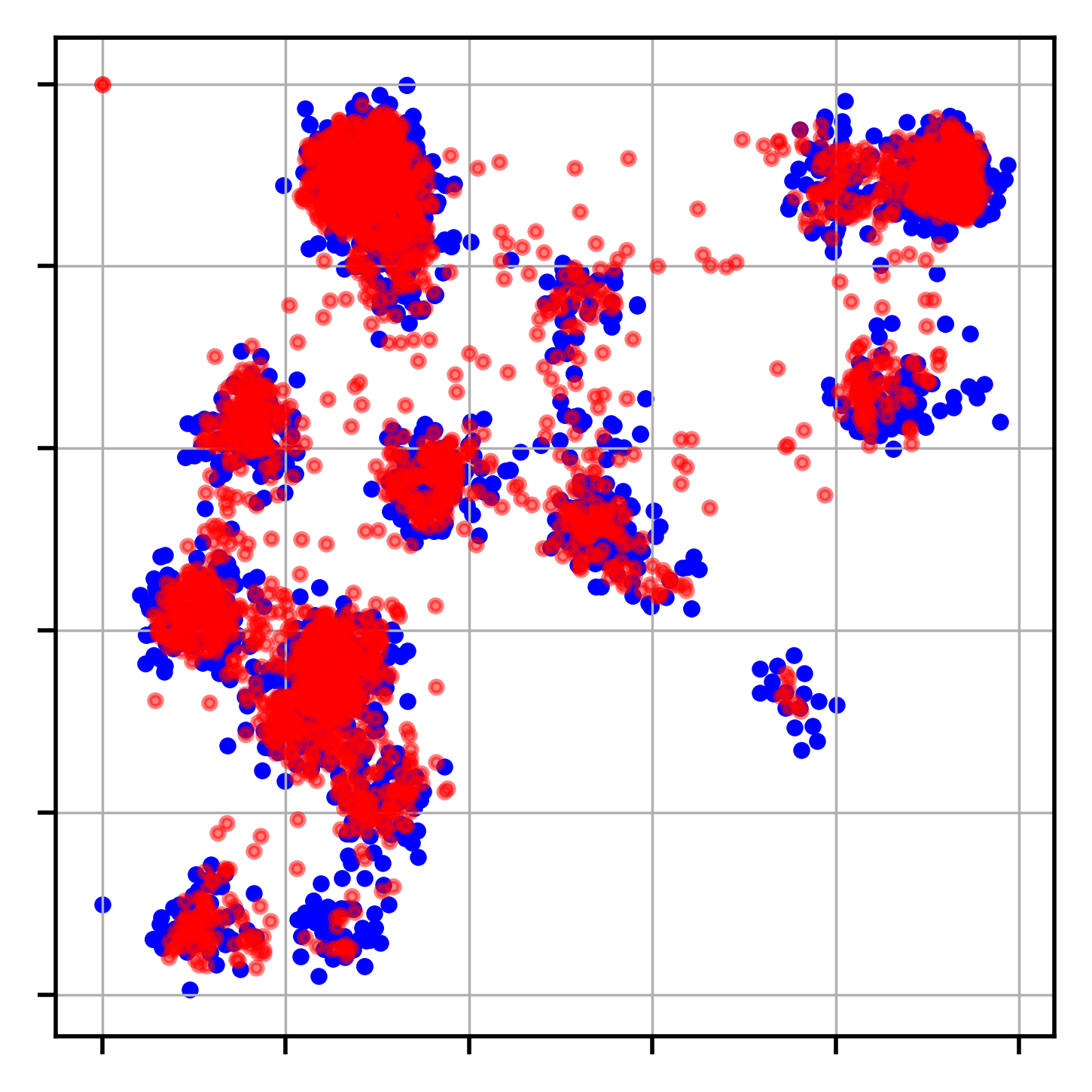}\vspace{-9pt}
\caption{\footnotesize Scalar}
\end{subfigure}
\begin{subfigure}{.24\textwidth}\includegraphics[width=.9\linewidth]{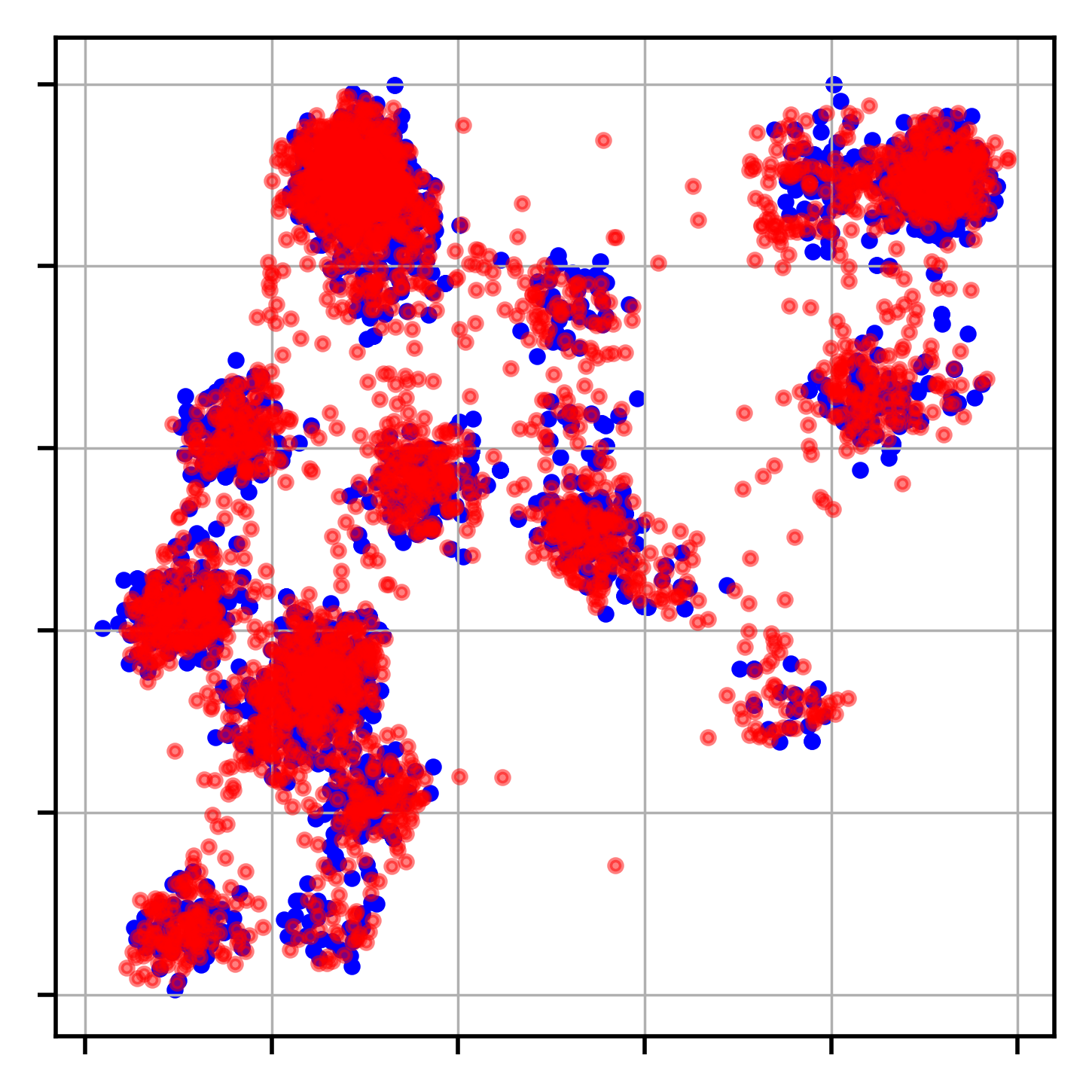}\vspace{-9pt}
\caption{\footnotesize Vector-Matrix}
\end{subfigure}
\begin{subfigure}{.24\textwidth}\includegraphics[width=.9\linewidth]{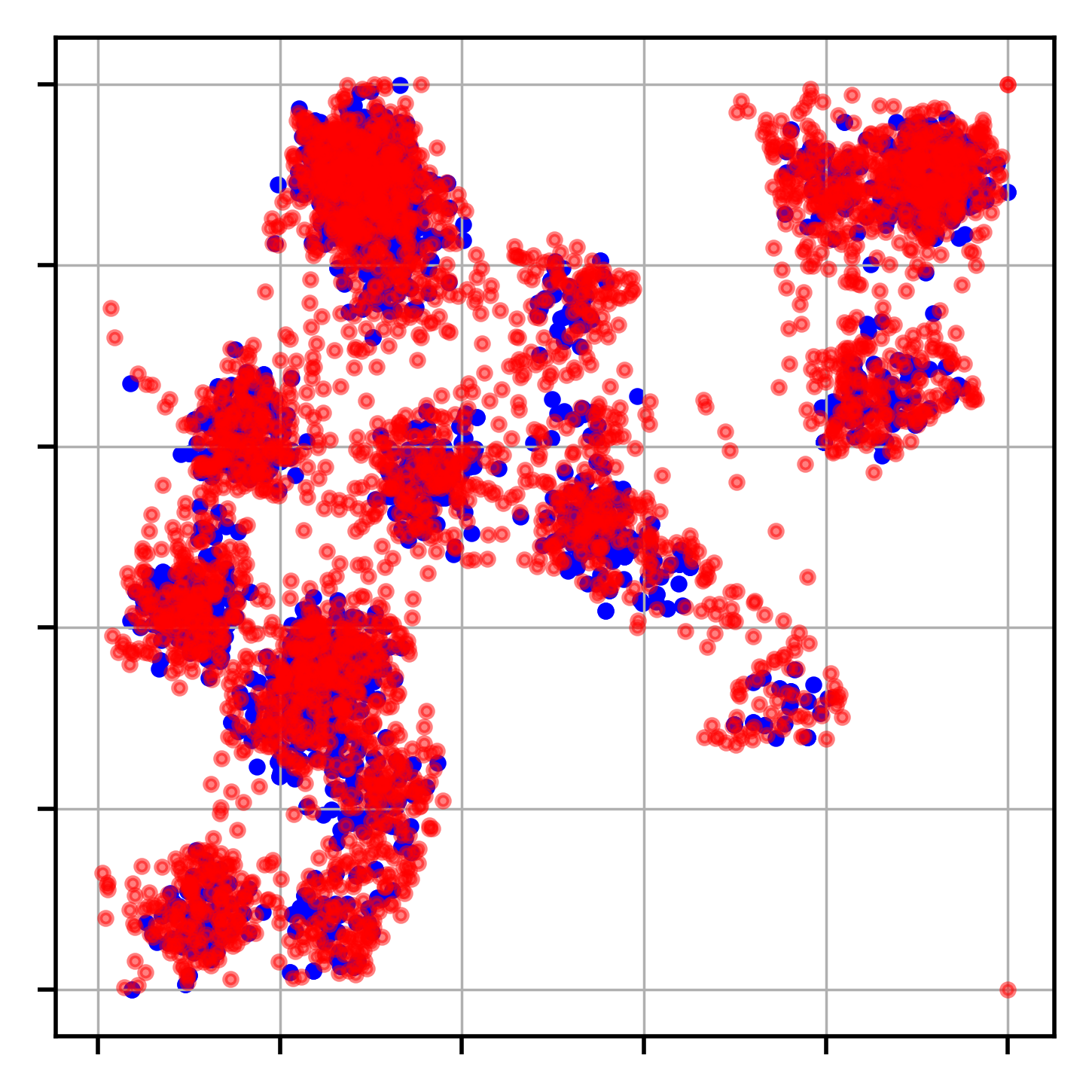}\vspace{-9pt}
\caption{\footnotesize Matrix-Matrix}
\end{subfigure}
\begin{subfigure}{.24\textwidth}\includegraphics[width=.9\linewidth]{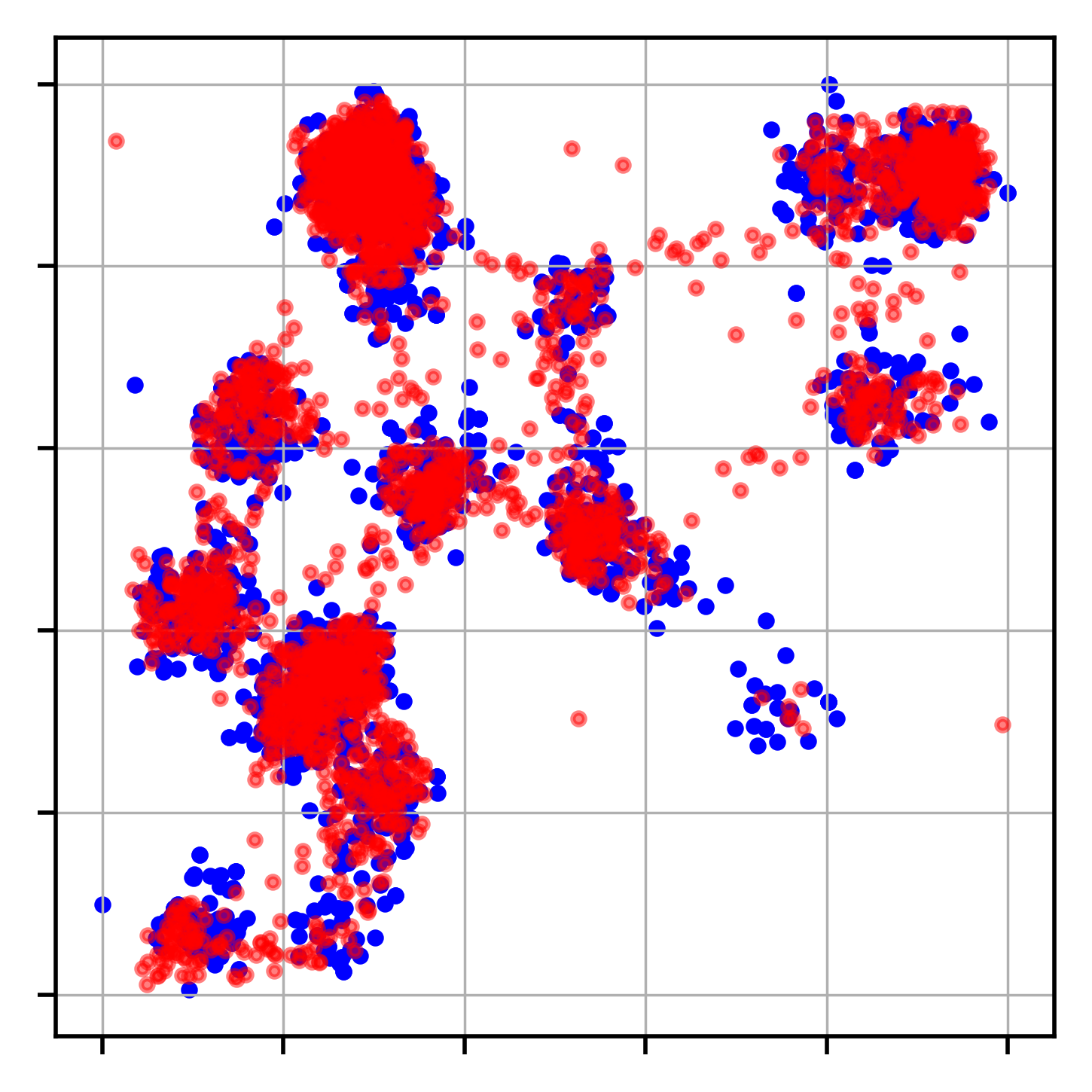}\vspace{-9pt}
\caption{\footnotesize SVD}
\end{subfigure}\vspace{-10pt}
\caption{\footnotesize Data samples (blue dots) and generated samples (red dots) with MDNs.}
\label{identity_2}
\end{figure}

The procedure is first sampling noise $u_1, u_2, \cdots, u_N$ from a prior, mapping them through a neural network to generate samples $X_1', X_2', \cdots, X_N'$, and then applying the costs between $X_1, X_2, \cdots, X_N$ and $X_1', X_2', \cdots, X_N'$. At each training step, it is applied to a different batch. Fig.~\ref{identity_2} shows fitting a $10$-state Gaussian mixture with randomly initialized means. The network's input is $10$D uniform noise. The variances $v$ in the costs are fixed at $0.001$.

% The procedure is first sampling noise $u_1, u_2, \cdots, u_N$ from a prior, mapping them through a neural network to generate samples $X_1', X_2', \cdots, X_N'$, and then applying the costs between $X_1, X_2, \cdots, X_N$ and $X_1', X_2', \cdots, X_N'$. At each training step, it is applied to a different batch. Figure~\ref{identity} shows the generated samples for a constructed $10$-state mixture density with randomly initialized means. The network's input is $10$-dimensional uniform noise. The variances $v$ in the costs are fixed at $0.001$.\vspace{-5pt}

\vspace{-5pt}

\begin{figure}[h]
\captionsetup[subfloat]{labelformat=parens, labelsep=space, font=small, skip=0pt} % Adjust the skip value as needed
\centering
\subfloat[\small Data density]{\includegraphics[width=0.32\textwidth]{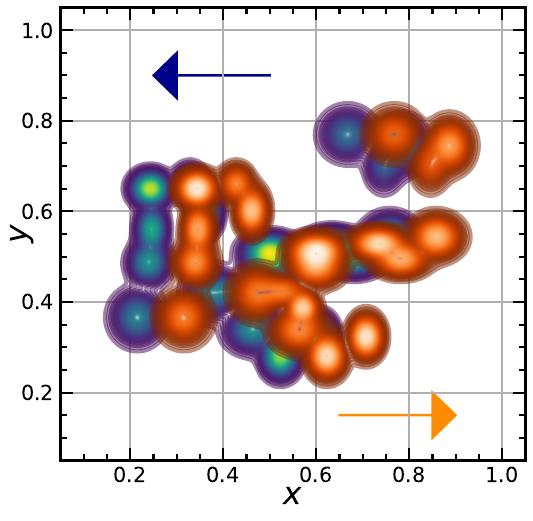}\label{shifting}}
\subfloat[\small Cost values]{\includegraphics[width=0.58\textwidth]{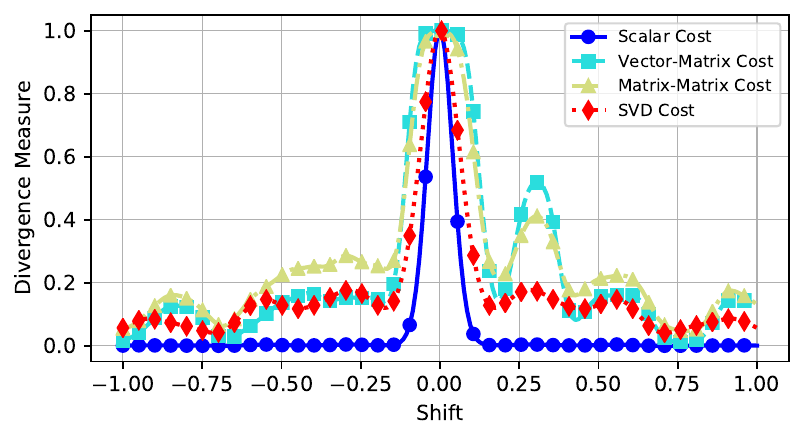}\label{CSD}} \vspace{-10pt}
% \caption{\footnotesize Comparisons of the values of costs by shifting the density $q$ away from $p$.}
\caption{\small Comparisons of cost value by shifting density $q$ away from $p$.\vspace{-11pt}}
\label{fig:csd_estimation}
\end{figure}

\noindent \textbf{\textit{As a measure.}} For densities $q$ and $p$, the closer they are, the larger the cost, as maximizing the cost will make $q$ approach $p$. We create two mixture densities in Fig.~\ref{shifting} and shift one away from the other starting with a distance of $-1$, moving $q$ towards $p$ until they match, and then away until the distance is $1$. We visualize scalar cost $sc$, vector-matrix cost $vc$, matrix-matrix cost $mc$, and SVD cost in Fig.~\ref{CSD}, normalizing them with peak values of $1$. We set $v=0.01$.

Comparing the four costs, the SVD cost is the most accurate descriptor. The vector-matrix cost and the matrix-matrix cost (we quantify $Trace(\mathbf{P}_{FG}^\intercal \mathbf{R}_F^{-1} \mathbf{P}_{FG})$) are not symmetrical as they use $q$ to predict $p$. The scalar cost may saturate much faster than the others.

\begin{figure}[h]
\centering
\begin{subfigure}{.22\textwidth}\includegraphics[width=\linewidth]{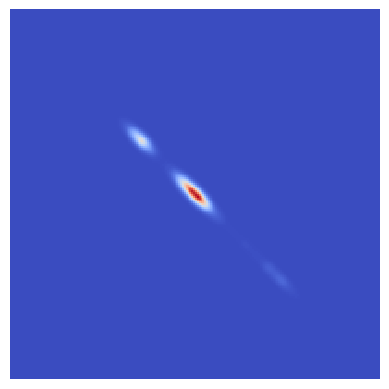}
\caption{$shift\;0$}
\end{subfigure}
\begin{subfigure}{.22\textwidth}\includegraphics[width=\linewidth]{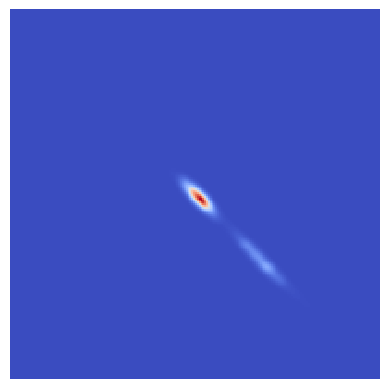}
\caption{$shift+0.5$}
\end{subfigure}
\begin{subfigure}{.22\textwidth}\includegraphics[width=\linewidth]{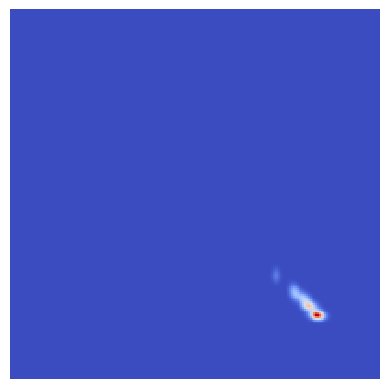}
\caption{$shift+1$}
\end{subfigure}
\begin{subfigure}{.22\textwidth}\includegraphics[width=\linewidth]{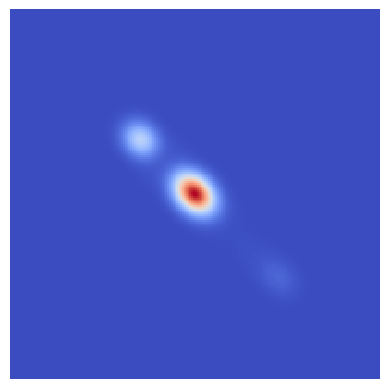}
\caption{$var=0.01$}
\end{subfigure}\vspace{-9pt}
\caption{\footnotesize Illustrating the SVD cost's property of approximating a diagonal function using Gaussian residuals (Eq.~\eqref{quantity}). (a)$\sim$(c): $var=0.01$. (d): $var=0.001$, which still approximates an identity function but less accurately.\vspace{-10pt}}
\label{identity}
\end{figure}

We have shown an important property of the SVD cost that it uses Gaussian residuals to find the best approximator for the identity function $\mathbbm{1}\{X=X'\}$. Fig.~\ref{identity} visualizes the quantity in Eq.~\eqref{quantity} and indeed it approximates an identity matrix. For this figure, we use only one dimension from Fig.~\ref{shifting} and pick $v=0.001$. As $q$ shifts away from $p$, the diagonal elements disappear. When choosing $v=0.01$, it still approximates a diagonal matrix but less accurately.\vspace{5pt}

\begin{figure}[h]
\centering
\begin{subfigure}{.45\textwidth}\includegraphics[width=\linewidth]{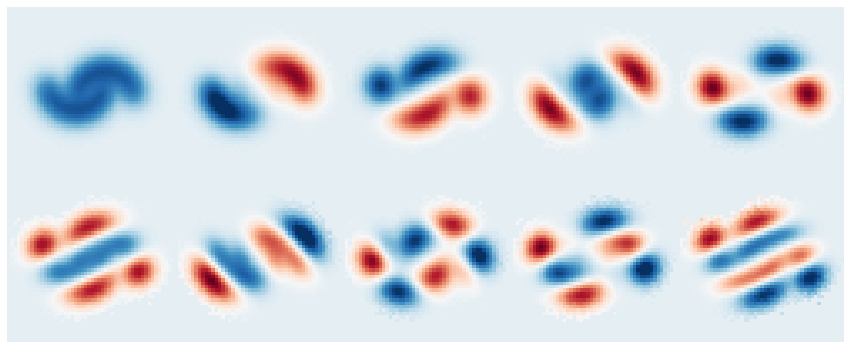}\vspace{-4pt}
\caption{\small Left singular function $q\neq p$}\label{3a}
\end{subfigure}\hspace{4pt}
\begin{subfigure}{.45\textwidth}\includegraphics[width=\linewidth]{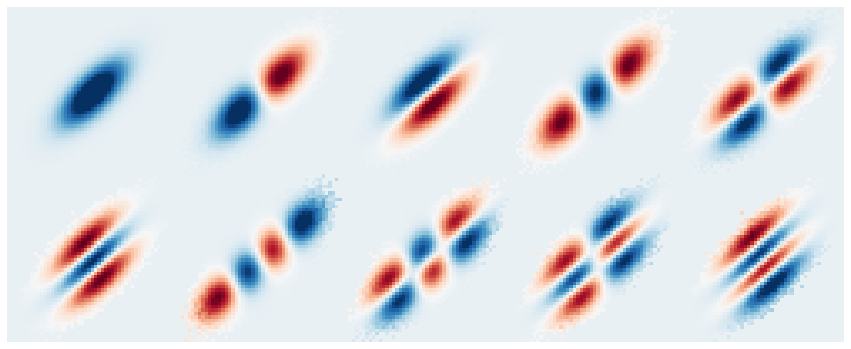}\vspace{-4pt}
\caption{\small Right singular function $q\neq p$}\label{3b}
\end{subfigure}\vspace{1pt}
\begin{subfigure}{.45\textwidth}\includegraphics[width=\linewidth]{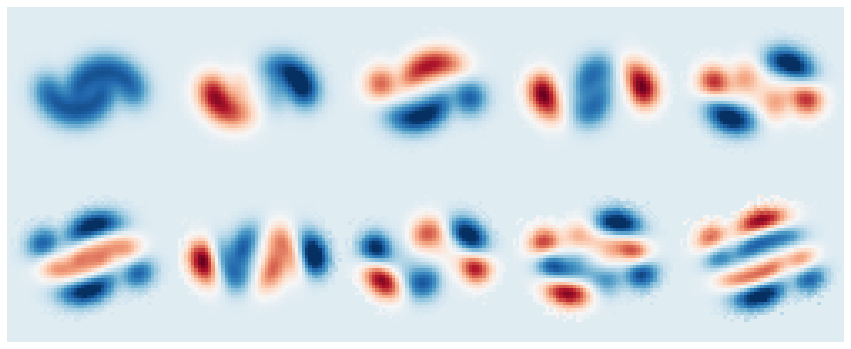}\vspace{-4pt}
\caption{\small Eigenfunction when $q = p$}\label{3c}
\end{subfigure}\vspace{-8pt}
\caption{\small Visualizations of singular functions for two-moon $q$ and Gaussian $p$, and eigenfunctions when $p$ and $q$ are both two moons.\vspace{-12pt}}
\end{figure}

\noindent \textbf{\textit{Visualizing the bases.}} Another important result we showed is that optimizing an SVD cost is decomposing the function $\sqrt{p(X)}\mathcal{K}(X,X')\sqrt{q(X)}$. Suppose $p(X)$ is a two-moon and $q(X)$ is a single Gaussian. Fig.~\ref{3a} and~\ref{3b} visualize the left and right singular functions. The shapes of them have an interesting consistency. And if $q$ and $p$ are both two-moon, the eigenfunctions have the shapes in Fig.~\ref{3c}. The singular functions of this decomposition do exhibit meaningful patterns.\vspace{3pt}

\noindent \textbf{\textit{Classification w/ multivariate approximator.}} In Sec~\ref{section_multivariate}, we extended $k(X,X') = \sum_{k=1}^K {\phi_k}(X){\psi_k}(X')$ to function approximator $k(X_1,X_2,\cdots,X_T)$ for series of variables, which first maps each pixel or patch to multivariate features, and features interact only through a Gaussian function in the final layer (Eq.~\eqref{equation1}~\eqref{equation2}~\eqref{equation3}). Each layer in the network defines centers of a mixture. We conducted experiments on MNIST and CIFAR10, comparing it to regular neural networks for classification. We found high accuracy even with pixel-level feature projections, implying that the relationship between the feature projections and pixel-level interactions may be separate. It also implies that factorizing multivariate functions by Gaussian products can be a good choice.\vspace{-7pt}

% \noindent \textbf{\textit{Classification w/ the multivariate approximator.}} We have proposed in Sec~\ref{section_multivariate} an extension from $k(X,X') = \sum_{k=1}^K {\phi_k}(X){\psi_k}(X')$ to a function approximator $k(X_1,X_2,\cdots,X_T)$ for a series of variables, which first maps each pixel or patch to multivariate features, and then only in the final layer do the features for pixels or patches interact through a Gaussian function (Eq.~\eqref{equation1}~\eqref{equation2}~\eqref{equation3}). We conduct experiments to show its use for MNIST and CIFAR10 classification, and compare it with a regular neural network. Interestingly, we found that even with feature projections at the pixel level, the accuracy is quite high. If the first layer of the network receives a patch instead of just pixels, the results are competitive with a regular network. This implies that the relationship between feature projections and the interactions at the pixel level may be separated. This also suggests that factorizing a multivariate function by the product of Gaussians can be a good choice.\vspace{-7pt}
\begin{table}[h] % Use [h!] to suggest placement "here" more strongly\
\caption{\small Classification using a series of the product of Gaussian functions as a function approximator (Eq.~\eqref{change_mixture_decoder}, Eq.~\eqref{equation1} to Eq.~\eqref{equation3}).\vspace{-10pt}}
\hspace{-6pt}\resizebox{.65\linewidth}{!}{
\footnotesize
\label{tab:model_accuracy}
\sisetup{table-format=1.3,      % Set a consistent format for numbers (e.g., X.XXX)
         output-decimal-marker={.}, % Use a period for the decimal marker
         table-space-text-post=\,\% } % Reserve space for the % sign in the header
\begin{tabular}{l S S S S}
    \toprule[2pt]
    \textbf{Model} & \multicolumn{2}{c}{\textbf{MNIST}} & \multicolumn{2}{c}{\textbf{CIFAR-10}} \\
    \cmidrule(lr){2-3} \cmidrule(lr){4-5} % Lines under the dataset names only
    \textbf{Patch Size} & \textbf{Train Acc} & \textbf{Test Acc} & \textbf{Train Acc} & \textbf{Test Acc} \\
    \midrule
    1 (Pixel-Level)& 0.990 & {0.974} & {0.899} & {0.600} \\
    3 & 0.998 & {0.982} & {0.997} & {0.806} \\
    5 & 1.000 & {0.988} & {0.998} & {0.819} \\
    7 & 1.000 & {0.989} & {0.998} & {0.820} \\
    \midrule
    CNN & 1.000 & {0.990} & {0.999} & {0.908} \\
    \bottomrule[2pt]
\end{tabular}}\vspace{-4pt}
\end{table}

% \section{Conclusion}

% \noindent \textbf{\textit{Visualize the bases. }}

% \noindent \textbf{\textit{Classification w/ the multivariate approximator. }}

% \newpage \;
% \newpage \;

\bibliography{reference}
\appendices

\end{document}